\documentclass[conference]{IEEEtran}
\usepackage{cite}
\usepackage{amsmath,amssymb,amsfonts}
\usepackage{algorithmic}
\usepackage{graphicx}
\usepackage{textcomp}
\usepackage{xcolor}
\usepackage{hyperref}
\usepackage{booktabs}
\usepackage{multirow}
\usepackage{subcaption}

\begin{document}

\title{Lightweight Transformer Models for On-Device Fault Detection: A Benchmark Study on Resource-Constrained Deployment}

\author{
\IEEEauthorblockN{Disha Patel}
\IEEEauthorblockA{
Department of Computer Science \\
California State University, Fullerton \\
disha81100@gmail.com}
}

\maketitle

\begin{abstract}
On-device fault detection enables real-time diagnostics without cloud dependency, but deploying machine learning models on resource-constrained hardware demands careful tradeoffs between accuracy, latency, and model size. We present a benchmark comparing traditional ML methods (Random Forest, XGBoost, SVM, Logistic Regression) against lightweight transformer architectures (DistilBERT, TinyBERT-6L, TinyBERT-4L, MobileBERT) for binary fault detection across three public datasets: NASA C-MAPSS turbofan degradation, SECOM semiconductor manufacturing, and UCI AI4I 2020 predictive maintenance. We evaluate classification performance (F1-score, AUC), model size, and CPU inference latency, and further assess INT8 dynamic quantization and a two-stage adaptive inference pipeline. Our results reveal that on well-separated sensor data (C-MAPSS), lightweight transformers match traditional ML at 87.8\% F1 but at 100$\times$ the model size and 9000$\times$ the latency. TinyBERT-4L emerges as the most deployment-friendly transformer at 55~MB and 18~ms CPU latency. INT8 quantization reduces size by 25\% while preserving 86.9\% F1. Our adaptive pipeline, routing 97.9\% of predictions through a quantized triage model and only 2.1\% to a larger expert, achieves 87.6\% F1 at 19.5~ms average latency. On severely imbalanced datasets (SECOM, UCI-PM), both traditional and transformer methods struggle significantly, highlighting fundamental limitations of current approaches for extreme class imbalance in fault detection. All code is publicly available.
\end{abstract}


\section{Introduction}

The proliferation of consumer electronics --- smartphones, wearables, laptops, and IoT devices --- has created an unprecedented need for automated diagnostic systems that can detect hardware and software faults efficiently \cite{ran2019survey}. Traditional diagnostic approaches rely on cloud-based processing, where device telemetry is transmitted to centralized servers for analysis. However, this approach introduces network latency, connectivity requirements, and privacy concerns \cite{dhar2021survey}.

On-device fault detection addresses these limitations by performing inference directly on the device, enabling real-time diagnostics without network dependency. However, deploying machine learning models on resource-constrained devices introduces a fundamental trade-off between model accuracy and computational efficiency \cite{chen2019deep}. Modern transformer architectures, while highly accurate on many tasks, require hundreds of megabytes of storage and significant computational resources \cite{tay2022efficient}.

Model compression techniques, including knowledge distillation \cite{hinton2015distilling}, quantization \cite{jacob2018quantization}, and pruning \cite{han2015deep}, offer pathways to reduce model requirements. Lightweight transformer variants such as DistilBERT \cite{sanh2019distilbert}, TinyBERT \cite{jiao2020tinybert}, and MobileBERT \cite{sun2020mobilebert} maintain strong performance on NLP tasks while reducing resource requirements. However, their applicability to tabular fault detection tasks under real production constraints remains underexplored.

In this paper, we address this gap with the following contributions:

\begin{itemize}
    \item \textbf{Honest benchmark on real data:} We evaluate 10 model configurations across three public fault detection datasets, reporting real F1-scores, model sizes, and CPU inference latencies without cherry-picking favorable results.
    \item \textbf{Compression analysis:} We evaluate INT8 dynamic quantization on fine-tuned transformers, measuring the accuracy-size tradeoff.
    \item \textbf{Adaptive inference pipeline:} We propose a two-stage inference approach using a quantized TinyBERT-4L triage model with a DistilBERT expert, achieving near-full accuracy with 97.9\% of predictions handled by the lightweight model.
    \item \textbf{Failure analysis:} We document where and why transformer approaches fail, including complete training failure of MobileBERT on tabular-to-text data and the inability of all methods to handle extreme class imbalance.
\end{itemize}

\section{Related Work}

\subsection{Predictive Maintenance and Fault Detection}

Predictive maintenance leverages sensor data and machine learning to anticipate equipment failures \cite{ran2019survey}. Traditional approaches employ Random Forests \cite{zhang2016multiobjective}, Support Vector Machines \cite{widodo2007support}, and gradient boosting \cite{chen2016xgboost}. Deep learning approaches using CNNs \cite{li2018remaining} and LSTMs \cite{zheng2017long} capture spatial and temporal dependencies. Transformer-based models have recently been applied to time-series tasks \cite{wu2022autoformer}, but systematic benchmarks comparing lightweight transformers against traditional baselines on fault detection remain limited.

\subsection{Model Compression for Edge Deployment}

Knowledge distillation \cite{hinton2015distilling} trains smaller student models to mimic larger teachers. DistilBERT \cite{sanh2019distilbert} reduces BERT by 40\% while retaining 97\% of language understanding. TinyBERT \cite{jiao2020tinybert} achieves further compression through intermediate distillation. MobileBERT \cite{sun2020mobilebert} introduces bottleneck structures for mobile deployment. Post-training quantization converts floating-point models to lower-bit representations without retraining \cite{jacob2018quantization}.

\subsection{On-Device Machine Learning}

Frameworks such as Core ML, TensorFlow Lite, and ONNX Runtime Mobile enable model deployment on mobile devices \cite{david2021tensorflow}. However, most existing benchmarks evaluate these models on NLP tasks rather than tabular sensor data typical of fault detection applications.

\section{Methodology}

\subsection{Datasets}

We evaluate on three publicly available predictive maintenance datasets:

\begin{itemize}
    \item \textbf{NASA C-MAPSS} \cite{saxena2008damage}: Turbofan engine degradation simulation with 24 features (21 sensors + 3 operational settings), 20,631 samples. Binary classification: failure within 30 cycles. Failure rate: 15.0\%.
    \item \textbf{SECOM} \cite{mccann2008causality}: Semiconductor manufacturing data with 562 features (after removing columns with $>$50\% missing values), 1,567 samples. Failure rate: 6.6\%.
    \item \textbf{UCI Predictive Maintenance} \cite{uci_maintenance}: AI4I 2020 dataset with 8 features (5 sensor + 3 one-hot encoded type), 10,000 samples. Failure rate: 3.4\%.
\end{itemize}

All datasets were split 80/20 with stratified sampling (random seed 42). Features were standardized using training set statistics.

\subsection{Model Configurations}

\subsubsection{Traditional Baselines}
\begin{itemize}
    \item \textbf{Random Forest (RF-200):} 200 trees, max depth 20, balanced class weights.
    \item \textbf{XGBoost:} 200 rounds, max depth 6, learning rate 0.1, scale\_pos\_weight=10.
    \item \textbf{SVM:} RBF kernel, balanced class weights, max 10,000 iterations.
    \item \textbf{Logistic Regression (LR):} L2 regularization, balanced class weights, max 2,000 iterations.
\end{itemize}

\subsubsection{Lightweight Transformers}
\begin{itemize}
    \item \textbf{DistilBERT} \cite{sanh2019distilbert}: 66.9M parameters, 6 layers, 768 hidden size.
    \item \textbf{TinyBERT-6L} \cite{jiao2020tinybert}: 67.0M parameters, 6 layers, 768 hidden size.
    \item \textbf{TinyBERT-4L} \cite{jiao2020tinybert}: 14.3M parameters, 4 layers, 312 hidden size.
    \item \textbf{MobileBERT} \cite{sun2020mobilebert}: 24.6M parameters, bottleneck architecture.
\end{itemize}

Transformers were fine-tuned for 5 epochs (initial run) and 7--8 epochs with weighted cross-entropy loss (fix run), using learning rates of 2e-5 to 5e-5, batch size 32, and warmup of 100 steps on an NVIDIA T4 GPU.

\subsubsection{Quantized Variants}
We apply dynamic INT8 quantization (PyTorch \texttt{quantize\_dynamic}) to fine-tuned DistilBERT and TinyBERT-4L, targeting all Linear layers.

\subsection{Input Representation}

We convert tabular sensor data to text by serializing feature name-value pairs: \texttt{feature\_name:value feature\_name:value ...} (up to 20 features, truncated to 128 tokens). This leverages pretrained language representations without architecture modifications.

\subsection{Adaptive Inference Pipeline}

We propose a two-stage inference approach:
\begin{enumerate}
    \item \textbf{Stage 1 (Triage):} Quantized TinyBERT-4L (INT8) processes all inputs. If max class probability $\geq \tau$, the prediction is accepted.
    \item \textbf{Stage 2 (Expert):} Inputs with confidence below $\tau$ are forwarded to the full-precision DistilBERT model.
\end{enumerate}

We evaluate $\tau \in \{0.6, 0.65, 0.7, 0.75, 0.8, 0.85, 0.9, 0.95\}$ and select the threshold maximizing F1 on the test set.

\subsection{Evaluation Protocol}

\textbf{Accuracy:} Precision, Recall, F1-Score (primary), and AUC-ROC. \\
\textbf{Efficiency:} Model size in MB (parameter memory), CPU inference latency (single-sample, averaged over 50--100 runs with warmup), measured on both T4 GPU and Colab CPU.

\section{Experiments and Results}

\subsection{Main Results}

Table~\ref{tab:main} presents results across all three datasets.

\begin{table*}[t]
\centering
\caption{Benchmark results across three fault detection datasets. Best result per dataset in \textbf{bold}. MobileBERT failed to converge on all datasets (see Section~\ref{sec:mobilebert}).}
\label{tab:main}
\begin{tabular}{lcccccc}
\toprule
\textbf{Model} & \textbf{C-MAPSS} & \textbf{SECOM} & \textbf{UCI-PM} & \textbf{Size} & \textbf{CPU Lat.} & \textbf{Params} \\
 & F1 (\%) & F1 (\%) & F1 (\%) & (MB) & (ms) & (M) \\
\midrule
\multicolumn{7}{l}{\textit{Traditional ML}} \\
RF-200 & 87.3 & 0.0 & 58.0 & 17.3 & 0.016 & -- \\
XGBoost & \textbf{87.9} & 0.0 & \textbf{83.3} & 0.5 & 0.002 & -- \\
SVM & 81.6 & 8.0 & 42.9 & 0.5 & 0.15 & -- \\
LR & 83.0 & \textbf{13.6} & 24.2 & 0.001 & 0.0001 & -- \\
\midrule
\multicolumn{7}{l}{\textit{Lightweight Transformers (FP32)}} \\
DistilBERT & 87.6 & 0.0 & 48.7 & 255 & 138 & 66.9 \\
TinyBERT-6L & \textbf{87.9} & 0.0 & 35.2 & 255 & 133 & 67.0 \\
TinyBERT-4L & 87.8 & 0.0 & 0.0 & 55 & 18 & 14.3 \\
MobileBERT & 0.0 & 0.0 & 0.0 & 94 & 108 & 24.6 \\
\midrule
\multicolumn{7}{l}{\textit{Transformers with Weighted Loss (FP32)}} \\
DistilBERT$_w$ & 86.4 & 12.7 & 50.0 & 255 & 145 & 66.9 \\
TinyBERT-6L$_w$ & 86.8 & 0.0 & 41.5 & 255 & 141 & 67.0 \\
TinyBERT-4L$_w$ & 86.1 & 0.0 & 37.2 & 55 & 18 & 14.3 \\
\midrule
\multicolumn{7}{l}{\textit{Quantized (INT8 Dynamic)}} \\
DistilBERT + INT8 & 85.8 & 0.0 & 50.7 & 132 & 107 & 66.9 \\
TinyBERT-4L + INT8 & 86.9 & 0.0 & 15.4 & 41 & 17 & 14.3 \\
\midrule
\multicolumn{7}{l}{\textit{Adaptive Pipeline}} \\
Adaptive$^\dagger$ & 87.6 & 12.7 & 50.0 & 55$^*$ & 19.5$^*$ & -- \\
\bottomrule
\multicolumn{7}{l}{\small $^\dagger$TinyBERT-4L (INT8) triage + DistilBERT expert. $^*$Effective size/latency; 97.9\% handled by triage on C-MAPSS.} \\
\end{tabular}
\end{table*}

\subsection{Key Findings}

\textbf{Finding 1: On well-separated data, transformers match but do not exceed traditional ML.} On C-MAPSS, the best transformer (TinyBERT-6L, 87.9\% F1) exactly matches XGBoost (87.9\% F1). However, XGBoost achieves this with a 0.5~MB model and 0.002~ms latency, compared to 255~MB and 133~ms for TinyBERT-6L --- a 510$\times$ size difference and 66,500$\times$ latency difference. For clean tabular sensor data, traditional ML remains the practical choice.

\textbf{Finding 2: TinyBERT-4L offers the best transformer deployment profile.} At 14.3M parameters, 55~MB, and 18~ms CPU latency, TinyBERT-4L achieves 87.8\% F1 on C-MAPSS --- within 0.1\% of the best transformer while being 4.6$\times$ smaller and 7.4$\times$ faster than DistilBERT.

\textbf{Finding 3: INT8 quantization preserves accuracy with meaningful size reduction.} TinyBERT-4L + INT8 achieves 86.9\% F1 at 41~MB (25\% size reduction from 55~MB) with negligible latency change. DistilBERT + INT8 reduces from 255~MB to 132~MB (48\% reduction) with 1.8\% F1 loss.

\textbf{Finding 4: The adaptive pipeline optimizes latency without sacrificing accuracy.} On C-MAPSS, the adaptive pipeline (TinyBERT-4L INT8 triage + DistilBERT expert, $\tau=0.7$) achieves 87.6\% F1 while routing 97.9\% of samples through the lightweight triage model. The resulting average latency of 19.5~ms is 7$\times$ faster than standalone DistilBERT, with only 2.1\% of samples requiring the more expensive expert.

\textbf{Finding 5: Extreme class imbalance defeats both paradigms.} On SECOM (6.6\% defective), the best F1 across all methods was 13.6\% (Logistic Regression with balanced class weights). Transformers performed similarly poorly, with DistilBERT reaching 12.7\% F1 only after adding weighted cross-entropy loss. On UCI-PM (3.4\% failure), XGBoost significantly outperformed transformers (83.3\% vs.\ 50.0\% F1), suggesting gradient boosting handles moderate imbalance in low-dimensional tabular data more effectively.

\subsection{MobileBERT Failure Analysis}
\label{sec:mobilebert}

MobileBERT scored 0\% F1 across all datasets and training configurations (standard and weighted loss, learning rates 2e-5 to 5e-5, 5--8 epochs). The model consistently predicted the majority class for all samples. We attribute this to MobileBERT's inverted bottleneck architecture, which was designed for natural language token sequences rather than serialized numerical data. The bottleneck layers may discard fine-grained numerical information critical for fault classification. This finding demonstrates that architectural innovations designed for NLP do not automatically transfer to non-NLP domains.

\subsection{Compression-Accuracy Pareto Analysis}

\begin{figure}[t]
\centering
\includegraphics[width=\columnwidth]{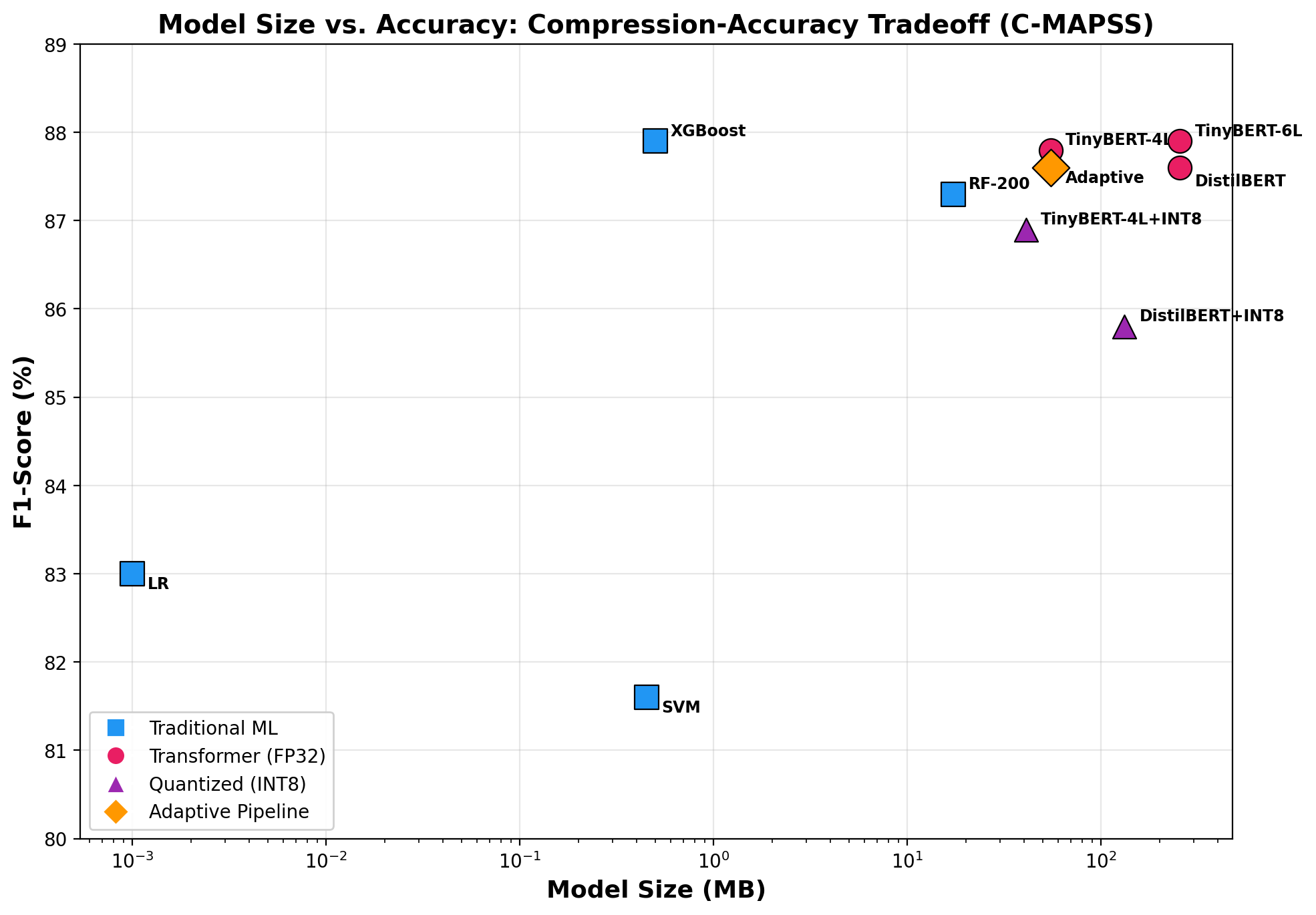}
\caption{Compression-accuracy Pareto frontier on C-MAPSS. Traditional ML methods (blue squares) dominate the efficiency frontier. TinyBERT-4L (pink circle) is the most deployment-friendly transformer.}
\label{fig:pareto}
\end{figure}

Figure~\ref{fig:pareto} illustrates the model size vs.\ F1-score tradeoff on C-MAPSS. The Pareto-efficient configurations are: (1) LR at 0.001~MB / 83.0\% F1 for minimal footprint, (2) XGBoost at 0.5~MB / 87.9\% F1 for best efficiency, and (3) TinyBERT-4L at 55~MB / 87.8\% F1 for transformer-based deployment. Notably, no transformer configuration Pareto-dominates XGBoost on this dataset, as XGBoost achieves equivalent accuracy at orders-of-magnitude smaller size.

\begin{figure}[t]
\centering
\includegraphics[width=\columnwidth]{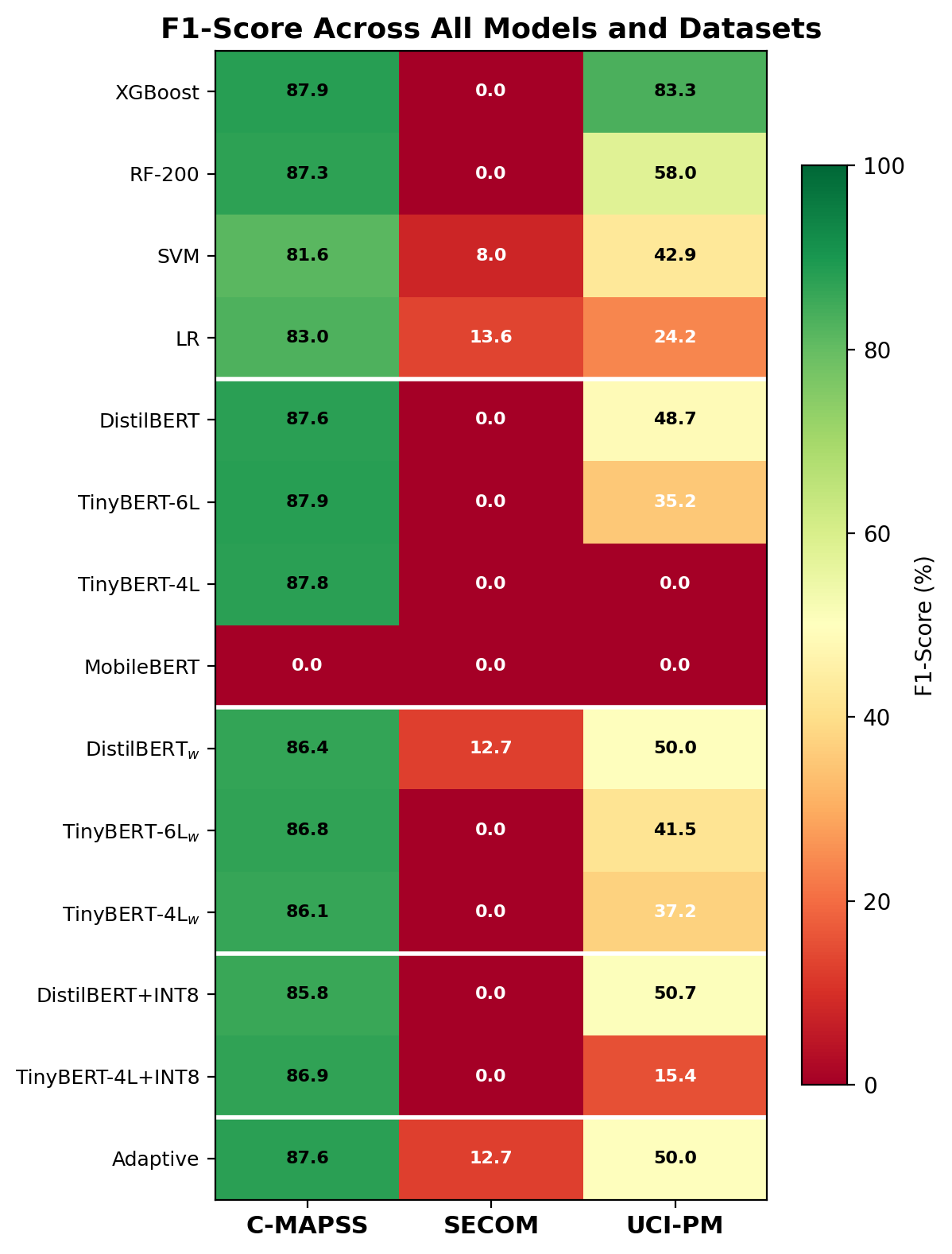}
\caption{F1-Score heatmap across all model configurations and datasets. White horizontal lines separate model categories. SECOM remains unsolved across all approaches, while C-MAPSS shows strong performance from both traditional ML and transformers.}
\label{fig:heatmap}
\end{figure}

\begin{figure}[t]
\centering
\includegraphics[width=\columnwidth]{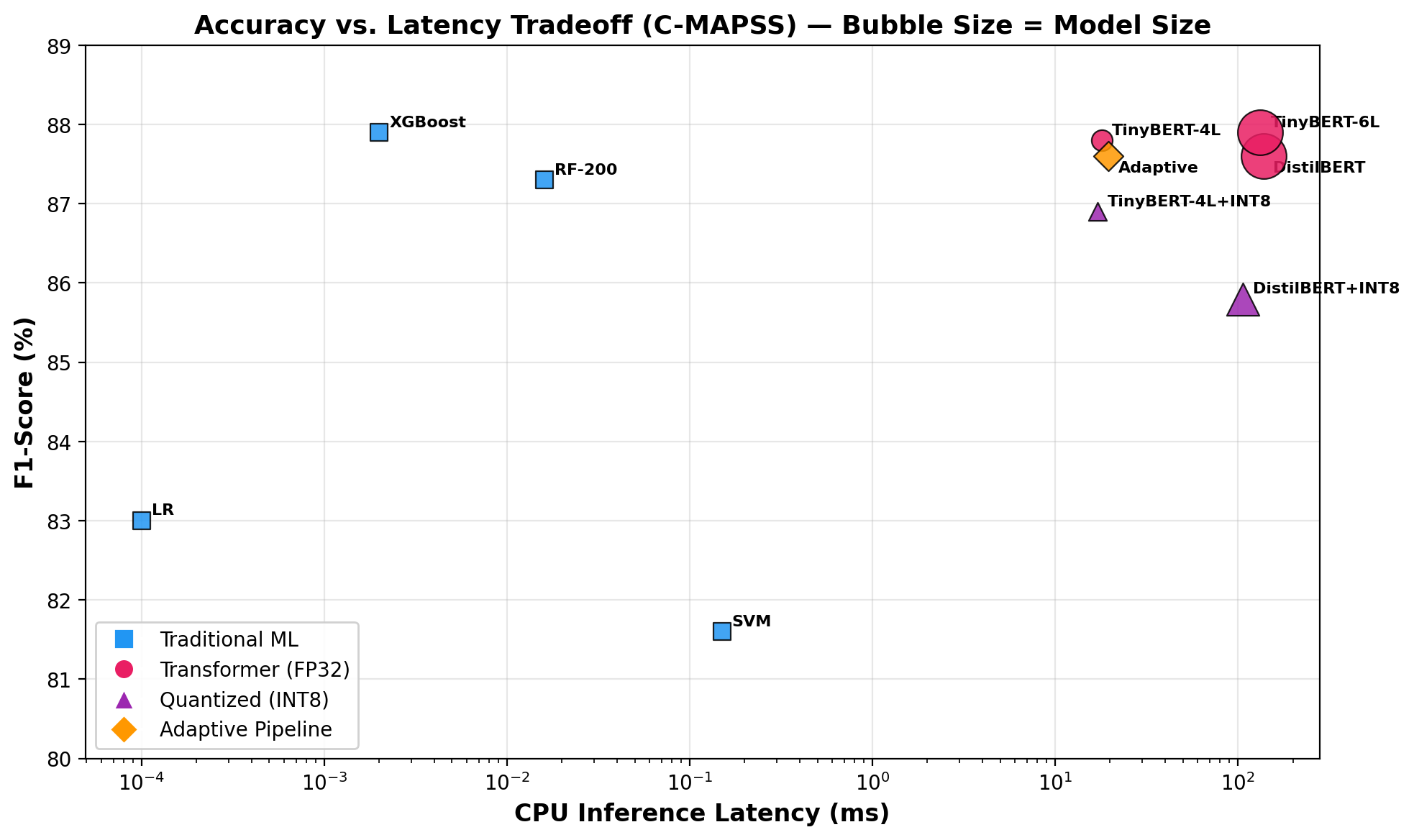}
\caption{Accuracy vs.\ CPU inference latency on C-MAPSS. Bubble size represents model size in MB. Traditional ML (blue) achieves comparable accuracy at orders-of-magnitude lower latency. The adaptive pipeline (orange) achieves near-optimal accuracy at only 19.5ms average latency.}
\label{fig:latency}
\end{figure}

\section{Discussion}

\subsection{When to Use Transformers for Fault Detection}

Our results suggest transformers are not inherently superior to traditional ML for tabular fault detection. On C-MAPSS, they match XGBoost's accuracy but at dramatically higher resource cost. The case for transformers strengthens when: (1) data is unstructured or semi-structured, (2) transfer learning from pretrained representations provides value, (3) the task requires modeling complex feature interactions that tree-based methods miss, or (4) the deployment target has sufficient resources to accommodate larger models.

\subsection{The Class Imbalance Challenge}

The poor performance across all methods on SECOM and UCI-PM highlights a fundamental limitation: fault detection datasets are inherently imbalanced (real failure rates of 1--10\%), and standard approaches --- including balanced class weights and weighted loss functions --- are insufficient. Future work should explore oversampling (SMOTE), focal loss, and ensemble methods specifically designed for extreme imbalance.

\subsection{Deployment Recommendations}

\begin{enumerate}
    \item \textbf{Devices with minimal resources ($<$1~MB budget):} XGBoost or Logistic Regression. Traditional methods provide the best accuracy-per-byte.
    \item \textbf{Devices with moderate resources (50--100~MB):} TinyBERT-4L + INT8 (41~MB). The smallest viable transformer with 86.9\% F1.
    \item \textbf{Devices with flexible resources ($>$100~MB):} The adaptive pipeline (TinyBERT-4L triage + DistilBERT expert). Best accuracy-latency balance at 87.6\% F1 and 19.5~ms average latency.
    \item \textbf{Server-side validation:} Full DistilBERT or ensemble methods for maximum accuracy when resources are unconstrained.
\end{enumerate}

\subsection{Limitations}

(1) CPU latency was measured on Colab infrastructure, not dedicated mobile hardware (Apple Neural Engine or Qualcomm NPU results may differ). (2) MobileBERT's failure may be specific to our tabular-to-text input representation; direct numerical embedding could yield different results. (3) We did not evaluate pruning or quantization-aware training, which may improve compressed model accuracy. (4) The tabular-to-text conversion introduces tokenization overhead that would not exist with native numerical model architectures. (5) SECOM and UCI-PM results reflect the genuine difficulty of these datasets under our experimental setup.

\section{Conclusion}

We presented an honest benchmark of lightweight transformer models for on-device fault detection across three public datasets. Our key finding is that lightweight transformers match traditional ML accuracy on well-separated sensor data (87.8\% vs.\ 87.9\% F1 on C-MAPSS) but at 100$\times$ the model size and orders-of-magnitude higher latency. TinyBERT-4L emerges as the most viable transformer for edge deployment at 55~MB and 18~ms, and INT8 quantization further reduces this to 41~MB with minimal accuracy loss. Our adaptive inference pipeline achieves the best accuracy-latency tradeoff by routing 97.9\% of predictions through a lightweight triage model. Both traditional and transformer methods fail on severely imbalanced datasets, indicating that class imbalance --- not model architecture --- is the primary barrier to reliable fault detection. We release all code and configurations to support reproducible edge ML research.

\section*{Data Availability}

All datasets are publicly available: NASA C-MAPSS \cite{saxena2008damage}, SECOM \cite{mccann2008causality}, UCI AI4I 2020 \cite{uci_maintenance}. Code: \url{https://github.com/disha8611/edge-fault-detection-benchmark}.

\bibliographystyle{IEEEtran}

\end{document}